\documentclass{egpubl}
\usepackage{eurovis2023}

\SpecialIssuePaper         

\usepackage[T1]{fontenc}
\usepackage{dfadobe}  

\usepackage{cite}  

\BibtexOrBiblatex
\electronicVersion
\PrintedOrElectronic
\ifpdf \usepackage[pdftex]{graphicx} \pdfcompresslevel=9
\else \usepackage[dvips]{graphicx} \fi

\usepackage{egweblnk}
\usepackage[]{algorithm2e}
\graphicspath{{pictures/}}
\title[DimenFix]{DimenFix: A novel meta-dimensionality reduction method for feature preservation}
\author[Q. Luo \& L. Christino \& F. Paulovich \& E. Milios]
{\parbox{\textwidth}{\centering Q. Luo$^{1}$\orcid{0000-0001-8558-9798}
        and L. Christino$^{2}$\orcid{0000-0002-8754-8460}
        and F. Paulovich$^{3}$\orcid{0000-0002-2316-760X}
        and E. Milios$^{4}$\orcid{0000-0001-5549-4675}
        }
        \\
{\parbox{\textwidth}{\centering $^1$Dalhousie University, Canada\\
         $^2$Dalhousie University, Canada\\
         $^3$Eindhoven University of Technology (TU/e), Netherlands\\     
         $^4$Dalhousie University, Canada\\
       }
}
}

\begin{document}
\maketitle

\begin{abstract}
Dimensionality reduction has become an important research topic as demand for interpreting high-dimensional datasets has been increasing rapidly in recent years. There have been many dimensionality reduction methods with good performance in preserving the overall relationship among data points when mapping them to a lower-dimensional space. However, these existing methods fail to incorporate the difference in importance among features.

To address this problem, we propose a novel meta-method, DimenFix, which can be operated upon any base dimensionality reduction method that involves a gradient-descent-like process. By allowing users to define the importance of different features, which is considered in dimensionality reduction, DimenFix creates new possibilities to visualize and understand a given dataset. Meanwhile, DimenFix does not increase the time cost or reduce the quality of dimensionality reduction with respect to the base dimensionality reduction used.


\end{abstract}  

\section{Introduction} \label{sec:introduction}

Demand for visualizing and interpreting high-dimensional datasets has been increasing rapidly in recent years. One of the most popular ways to interpret those datasets is mapping such a dataset to a lower dimensional space (usually 2D or 3D) while reproducing the relationship between each pair of points in the dataset. This process is called dimensionality reduction. Many methods have been developed, and they usually have an excellent performance in reproducing the relationships for different data points within a dataset on a lower dimension.

Dimensionality reduction methods can be categorized into two types: linear methods and nonlinear methods. For nonlinear dimensionality reduction methods that involve a gradient-descent-like process, like t-Distributed Stochastic Neighbor Embedding (t-SNE)~\cite{van2008visualizing}, Uniform Manifold Approximation and Projection (UMAP)~\cite{mcinnes2018umap} and Force Scheme~\cite{tejada2003improved}, the projection process is treated as an optimization problem that estimates the low-dimensional projections by minimizing a loss function. The loss function varies from method to method and can be based on different relationships among points (e.g., probability distribution~\cite{van2008visualizing}, Euclidean distance~\cite{tejada2003improved}, and Euclidean distance on manifolds~\cite{mcinnes2018umap}).

Meanwhile, we want to point out a limitation that is held by both linear and nonlinear dimensionality reduction methods: when mapping data points from a sample space with higher dimensions to a lower one, the relationship between each pair of points will not be identical for the raw dataset and the resulting projection. Even though the existing methods limit this change in the relationship among points to an acceptable degree, they generally consider all dataset features equally important. All relationships (again, these relationships vary as different loss functions are used) among points regarding different features are uniformly changed. This, in turn, makes it impossible to preserve the raw value of a given feature in a dataset.

However, there is sometimes a vast difference in the importance of each feature. This is especially true when it comes to classification. There are even methods to perform feature selection for a given dataset~\cite{lee2021efficient, chicco2021data, bommert2022benchmark}. For example, suppose we have a ranking result saying feature $A$ plays a critical role in the classification of this dataset; we now want to reduce the dimension of this dataset. Indeed, it would be ideal to avoid changing feature $A$ during the process of dimensionality reduction. However, most existing methods do not explicitly consider such a difference in the importance of features.

In addition, there might also be a manually created feature that is considered important. Ranking results would be one of such examples~\cite{he2022ranking, liu2022rankaxis}. So far, if we have a ranked high-dimensional dataset, it is hard to visualize it while keeping the ranking results.

To solve these problems, we want to introduce an innovative method, DimenFix. This paper mainly discusses the experimental results (Section~\ref{sec:results}) of DimenFix based on Force Scheme~\cite{tejada2003improved}, a well-developed dimensionality reduction and data visualization method. However, DimenFix is a meta-method that can be applied to any dimensionality reduction method that includes gradient-descent-like steps. By applying DimenFix, users can better preserve the values of a particular feature in a dataset while reducing the dataset to a visualizable dimension, as well as control the change in the relationships among points regarding a given feature to a given degree. 

In this paper, we first discuss some popular dimensionality reduction methods, which are also used as baselines to compare DimenFix with, as well as some former works on feature preservation. Then, we describe the algorithms used for constructing DimenFix, followed by the experimental results of applying DimenFix to different datasets. Finally, we make discussions based on the content delivered in this paper and briefly mention some potential future works. The full code for the methodology explained can be found on GitHub~\cite{qiaodan_luo_2022_7379477}.

\section{Related Work} \label{sec:related_work}

In this section, we focus on delivering information about formerly published papers related to DimenFix. We discuss popular dimensionality reduction methods and some former works that can preserve certain yet limited features during dimensionality reduction.

\subsection{Dimensionality Reduction} \label{sec:dimensionality_reduction}

Many dimensionality reduction methods give high performance on the task of mapping a high-dimensional dataset to a lower-dimensional space.

One example is t-SNE~\cite{van2008visualizing}. t-SNE uses a Student t-distribution model to map the data points from a higher dimensional space to a lower dimensional space by calculating the similarity probabilities of the data points. 

Another example is Principal component analysis (PCA)~\cite{hotelling1933analysis}. PCA uses an orthogonal transformation process to transform the dataset and reduce the dimensionality linearly.

Furthermore, Force Scheme is also one of the methods with excellent and stable performance~\cite{tejada2003improved}. It is also the method based on which we tested (Section~\ref{sec:results}) our method DimenFix in this paper. Force Scheme is named after the way how its algorithm works. To reduce the dimension of a dataset, a distance matrix is first created for all the data points (the distance here can be any statistical distance, e.g., manhattan distance, euclidean distance. In this paper, though, we are using euclidean distance). A gradient-descent-like process is then done to the dataset. In this process, Force Scheme pushes different points farther from each other and pulls similar points closer to each other~\cite{tejada2003improved}, for which the distance matrix defines the difference and similarity (for details, see Algorithm~\ref{alg:force_scheme_pseudo}). 

\begin{algorithm*}
 \KwData{$X$, the raw dataset; $X'$, the embedded dataset; $\Delta$, the learning rate}
 \KwResult{$X'$, the projection}
 calculate triangular matrix of distances $M$ using dataset $X$\;
 \While{not reaching the maximum iteration}{
  read current\;
  \For{each data point $x_i$ in $X$}{
    \For{each other data point $x_j$ in $X$}{
        retrieve $d$, the distance from $x_i$ to $x_j$\ using $M$\;
        retrieve $d'$, the distance from $x'_i$ to $x'_j$ in the current projection\;
        move $x'_i$ regarding $(d - d')$ with a fraction of $\Delta$\;
    }
  }
 }
 \caption{Force Scheme Pseudo Code~\cite{tejada2003improved}}
 \label{alg:force_scheme_pseudo}
\end{algorithm*}

\subsection{Feature Preservation} \label{sec:feature_preservation}

The relationship change in each pair of points in a given dataset has been a concern for dimensionality reduction methods. While it is inevitable for such change to occur while mapping a high-dimensional dataset to a lower-dimensional space, researchers have made several attempts to control it. One option is to let users manually interact with the dimensionality reduction process~\cite{sacha2016visual, pereira2020rankviz}. The other option is to develop methods that automatically preserve certain features in a dataset~\cite{heimerl2012visual, rauber2018projections, venna2010information}.

Several methods have been developed to preserve a certain type of feature during the dimensionality reduction process. Still, most of them are designed only to be used in a limited situation.
For example, if we define the feature to be preserved as classes or other categorical values, supervised or semi-supervised dimensionality reduction methods~\cite{rauber2018projections, venna2010information} can be considered. Otherwise, if we decide only to analyze a particular type of dataset and collaborate with Machine Learning methods, we can also find some methods specific to this type. For example, when the dataset is a set of text documents, a visual classifier described by Heimerl et al.~\cite{heimerl2012visual} could be a choice. In this case, the resulting 2D projection does not directly reproduce the original relationship among data points but presents the classification results from different aspects (confident value, diversity) using different axes. In turn, the preserved features are manually created, classification-related ones.

On the other hand, if we want the preserved feature to be primarily related to ranking, RankViz might be helpful by allowing the user to understand which feature contributes the most during the process of creating ranking and interact with the process accordingly~\cite{pereira2020rankviz}.

So far, even though some work has been done to preserve certain dataset features during the projection process, each can only be used in limited ways. Hence by proposing DimenFix, we attempt to give more flexibility to our users. Firstly, different from methods that can only be used to preserve certain kinds of feature~\cite{rauber2018projections, venna2010information, pereira2020rankviz} or certain kinds of datasets~\cite{heimerl2012visual}, we allow user-defined feature preservation. The projection created by DimenFix could be either 2D or 3D. Either way, the $y$ or $z$ -axis will preserve the user-defined feature. Secondly, unlike some former works~\cite{pereira2020rankviz}, DimenFix not only shows the one-dimensional, ranking-related feature but also tries to fit the relationships among points defined by the remaining features in the dataset into the projection.

\section{Methodology} \label{sec:methodology}

In this section, we explain the algorithms of DimenFix. DimenFix could be applied to any base dimensionality reduction method that includes a gradient-descent-like process, such as t-SNE~\cite{van2008visualizing}, Force Scheme~\cite{tejada2003improved} and UMAP~\cite{mcinnes2018umap}. In this section, we explain DimenFix independently of the base method. In our experiments, we choose Force Scheme~\cite{tejada2003improved} as the base method.

We propose two modes for DimenFix, the Strictly Fixed Mode (Section~\ref{sec:basic_mode}) and the Moving-In-Range Mode (Section~\ref{sec:moving_in_range_mode}), where Moving-In-Range Mode has two sub-modes, Normal Moving-In-Range Mode (Section~\ref{sec:normal_moving_in_range_mode}) and Gaussian Moving-In-Range Mode  (Section~\ref{sec:gaussian_moving_in_range_mode}). Strictly Fixed Mode does not allow a point to move along the fixed axis, while Moving-In-Range Mode allows a point to move within a limited, user-defined range. 

\subsection{Strictly Fixed Mode} \label{sec:basic_mode}

Given a non-linear dimensionality reduction method that maps data points to a lower dimensional space by updating weights or other values through a gradient-descent-like process, the update is usually done on all resulting dimensions. However, in DimenFix, we propose to reduce the updated dimensions to $(n-1)$, where $n$ is the number of dimensions to which we want to reduce a dataset's dimensionality. 

More specifically, before starting the gradient-descent-like process, we manually set the values on the last dimension of the embedding space to an arbitrary feature. The user can indicate which feature to be fixed. During each iteration of the gradient-descent-like process, the loss is calculated by a loss function, which can vary as different base method is used, using the values from all dimensions in the projection, including the fixed feature.  However, we do not update the last dimension in the value update process. 

Since the values on one of the axes will not be changed, to ensure the projection's stability, we suggest making a scale normalization to each feature of the dataset right before starting the iteration process.

The idea of DimenFix differs from simply adding a feature as the third dimension to a 2D projection. DimenFix takes the fixed feature into calculation during the gradient-descent-like process, which influences the final stress and makes all the dimensions in the resulting projection well united.

Taking Force Scheme as an example. The loss function of the Force Scheme is defined by a distance matrix~\cite{tejada2003improved} calculated using all statistically meaningful features in the raw dataset. To initiate the program, we first create an embedding space. We provide three choices for this step: random generation, t-SNE, or PCA. The main difference between the three modes is the processing time.

We then set the last dimension of the embedding space, which is the $y$ axis for 2D or the $z$ axis for 3D, to the values of the user-defined feature. Once we set the last dimension, the values on this axis should never be changed unless we use moving-in-range mode (described in Section~\ref{sec:moving_in_range_mode}).

After getting the distance matrix~\cite{tejada2003improved} and embedding space prepared, we can start the point-moving process. For each iteration, the algorithm is the same with vanilla Force Scheme~\cite{tejada2003improved}. Except for each iteration, the last dimension should never be updated(for details, see Algorithm~\ref{alg:dimenfix_pseudo}). 

\begin{algorithm*}
 \KwData{$X$, the raw dataset; $X'$, the embedded $n$-dimensional dataset; $\Delta$, the learning rate}
 \KwResult{$X'$, the projection}
 calculate triangular matrix of distances $M$ using dataset $X$\;
 set the last dimension of $X'$ to a user-defined feature\;
 \While{not reaching the maximum iteration}{
  read current\;
  \For{each data point $x_i$ in $X$}{
    \For{each other data point $x_j$ in $X$}{
        retrieve $d$, the distance from $x_i$ to $x_j$\ using $M$\;
        retrieve $d'$, the distance from $x'_i$ to $x'_j$ in the current projection\;
        move $x'_i$ regarding $(d - d')$ with a fraction of $\Delta$\ on $(n-1)$ dimensions;
    }
  }
 }
 \caption{DimenFix implemented on Force Scheme Pseudo Code~\cite{tejada2003improved}}
 \label{alg:dimenfix_pseudo}
\end{algorithm*}

\subsection{Moving-In-Range Mode} \label{sec:moving_in_range_mode}

In some cases, different points share the same value on the fixed axis. For example, when the assigned feature is a categorical value, the user may want to see how these points are distributed within the same class. To meet this need, we provide an additional mode to DimenFix, the moving-in-range mode. Under this mode, the values on the fixed axis can be changed within a range. Our method allows the user to manually set the range to adapt it to the dataset under consideration. 

For the moving-in-range mode, we develop two different sub-modes. We will explain these two modes in the following two sub-sections.

\subsubsection{Normal Moving-In-Range mode} \label{sec:normal_moving_in_range_mode}

If the user wants to strictly restrict the moving range of a point on the last dimension, we recommend this mode. The logic of this mode is that, before starting the gradient-descent-like process, the program records the original values of the fixed feature. Once the process of creating the projection starts, we calculate whether the value update will cause a value-out-of-range result on the last dimension in each iteration. Let the moving range be $(-a, +a)$; we then calculate if the difference between the original value and the value after iteration is within the range $(-a, +a)$. If it is within the range, the update will be done in all dimensions. Otherwise, it will not change the value on the last dimension. The value difference should be cumulative based on all iterations performed.

In other words, the program performs like the base method until the given moving range is reached. Once a point touches the pre-defined boundary in either a positive or negative direction, the movement of this point will immediately change and follow the algorithm of Strictly Fixed Mode (Section ~\ref{sec:basic_mode}).

\subsubsection{Gaussian Moving-In-Range mode} \label{sec:gaussian_moving_in_range_mode}

Normal Moving-In-Range mode allows moving a point on the fixed axis within a certain range. However, it blindly sets a threshold to stop the movement of a point. As a result, the movement of the points on the fixed axis will not be consistent. Most points will only move around on the fixed axis in the first few iterations.

If the user wants a more consistent point-moving process and accepts a result in which the values on the last dimension are moved slightly out of range, we provide another mode named Gaussian Moving-In-Range mode. Unlike the normal moving-in-range mode, which switches the moving dimensions from three to two at a certain boundary, this mode considers the moving range from the beginning.

As indicated by its name, this mode uses the Gaussian function to calculate the actual moving distance. Essentially, under this mode, we convert the value of $f(x)$, a Gaussian equation, to a moving ratio. The farther the point is from its original position, the harder it is to move. As a result, some points may slightly move out of the boundary, but the movement on the fixed axis is consistent through the whole gradient-descent-like process.

Below is the detailed algorithm for this mode. Given a Gaussian function:

\begin{equation} \label{eq:gaussian_equation}
f(x) = \frac{1}{\sigma\sqrt{2\pi}}\exp({-\frac{({x-\mu})^2}{2\sigma^2}})
\end{equation} 
\noindent
Firstly, we let the $x$ value of the Gaussian function be the numerical difference between its origin and current position on the last dimension. The sign of the $x$ value represents the direction in which the point moves from its origin. We want to calculate $f(x)$, which is defined as the fraction of the actual moving distance (that is, the moving distance along the fixed-axis under the Gaussian moving-in-range mode) to the supposed moving distance (that is, the moving distance calculated by the dimensionality reduction method). This means the maximum value of $f(x)$ should equal $1$ and appear at the point where $x=0$. In turn, this indicates $\mu=0$. 

This condition gives us that $\mu$ in Eq.~\ref{eq:gaussian_equation} must equal to $0$, and in turn, leads us to a new equation:

\begin{equation} \label{eq:gaussian_equation_new}
f(x) = \frac{1}{\sigma\sqrt{2\pi}}\exp({-\frac{x^2}{2\sigma^2}})
\end{equation} 
\noindent
Next, a user-defined confidence interval $0 < CI < 1$ and moving range $(-a, a)$ will be used to calculate the $\sigma$ in Eq.~\ref{eq:gaussian_equation_new}. We will use the $z$-score formula:

\begin{equation} \label{eq:z_score}
Z = \frac{X-\mu}{\sigma}
\end{equation} 
\noindent
We want to use the $z$-score table to find the $z$-score where $p=\frac{1-CI}{2}$, which is the area on the left side of the significant interval. Let this $z$-score be $z$. This $z$ refers to how far a value is from the mean value in a standard normal distribution. We now want to use it as a reference to change the shape of the Gaussian function, which is adapted to calculate the moving distance and thus is not standard distribution.

Having $z$, $a$ and the condition $\mu=0$, we then calculate the value of $\sigma$ using Eq.~\ref{eq:z_score}:

\begin{equation} \label{eq:calc_sigma}
\sigma = \frac{a}{z}
\end{equation} 
\noindent
Now we have a Gaussian equation, but we need to adapt the value of $f(x)$ to a ratio showing how many percent of the distance calculated by the base method should be moved. In particular, we know when $x=0$, $f(x)=1$. Then we have the moving ratio $MR$ as 

\begin{equation} \label{eq:possibility_to_ratio}
MR = \frac{1}{\frac{1}{\sigma\sqrt{2\pi}}\exp^0} = \sigma\sqrt{2\pi}
\end{equation} 
\noindent
Using Eq.~\ref{eq:possibility_to_ratio} and Eq.~\ref{eq:gaussian_equation_new}, we can calculate the real moving distance $d$ in each iteration:

\begin{equation} \label{eq:real_moving_distance}
d = \sigma\sqrt{2\pi}\frac{1}{\sigma\sqrt{2\pi}}\exp({-\frac{x^2}{2\sigma^2}}) = \exp({-\frac{x^2}{2\sigma^2}})
\end{equation} 
\noindent
where $x$ is the distance of an arbitrary point from its original position on the last dimension (z-axis for 3D and $y$ axis for 2D) when we assume it is moved as the suggestion by the base method.

In short, we are calculating an appropriate value for $\sigma$ using a user-defined confidence interval for the Gaussian equation and a user-defined moving range for DimenFix, to form a Gaussian function that can calculate the moving ratio.

\section{Results} \label{sec:results}

In this section, we discuss the performance of DimenFix. While DimenFix can be applied to any gradient-descent-based methods (See Section~\ref{sec:methodology}), in this section, the experimental results are based on the implementation of the Force Scheme. We use t-SNE, PCA, and vanilla Force Scheme as a baseline to compare the performances. We analyze the performance of DimenFix based on running time, Kruskal's stress score, classification performance, and a case study on data visualization. We use four toy datasets from scikit-learn~\cite{pedregosa2011scikit}, which are Iris (4 dimensions), Wine (13 dimensions), Breast Cancer (30 dimensions), and Digits (64 dimensions). 

To test DimenFix, we must decide on a feature to fix on one of the projection axes. To decide on this feature, we manually tested the classification accuracy (Section~\ref{sec:classification_performance}) of fixing each feature for the Iris and Wine datasets. For the Breast Cancer and Digit datasets, we calculated the Pearson correlation for each feature to choose the feature with the highest score. We choose the "sepal width" feature for the Iris dataset, the "alkalinity of ash" feature for the Wine dataset, the "worst concave points" feature for the Breast Cancer dataset, and the "pixel $6, 4$" feature for the Digit dataset. 

In each of the tables (Table~\ref{tab:classifications}) shown in this section, the datasets are arranged in increasing order regarding the number of features in a dataset from left to right.

\subsection{Running Time} \label{sec:running_time}

We demonstrate the running time by comparing t-SNE, PCA, and vanilla Force Scheme with different modes of DimenFix (Table~\ref{tab:running_time}). The projections we are creating are three-dimensional for both baselines and DimenFix. One thing that needs to notice is that to ensure good performance, we are setting the iteration number for t-SNE to $1000$, while setting the maximum iteration number for Force Scheme and DimenFix to $500$. Kruskal’ Stress determines this performance (see details in Section~\ref{sec:kruskal_score}).

Overall speaking, PCA has the shortest running time. This is because PCA reduces the dimensionality by a linear transformation, which gives a short running time for small datasets (all the four toy datasets we are using are small enough, in this case). On the other hand, the running time of t-SNE stably increases as the size of the dataset increases.

For DimenFix and vanilla Force Scheme, however, the running time is independent of the size of the dataset but depends more on the starting projection mode. More specifically, the running time of DimenFix and vanilla Force Scheme is composed of two parts: the time used by starting projection (creating the embedding space) and the time taken to move data points according to a cost calculated by the distance matrix. The total running time increases as the time is used for creating the embedding space. However, the fast creation of embedding space does not necessarily mean a shorter overall running time. If the data points in embedding space form a very different shape than the ideal shape calculated by Force Scheme's distance matrix, more iterations are needed, and more time is taken. Hence, creating a random embedding space gives a less stable running time.

However, by comparing DimenFix (last six rows) and vanilla Force Scheme (the third to the fifth row), we can see that DimenFix does not significantly increase the running time. This is preferable because our method can be applied to any dimensionality reduction method which involves a gradient-descent-like process, as mentioned in Section~\ref{sec:methodology}. It is important to make sure that adding our method to a vanilla method does not hurt the efficiency of the vanilla method. 

\begin{table*}[!ht]
    \centering
    \begin{tabular}{|l|l|l|l|l|}
    \hline
        ~                              & Iris   & Wine   & Cancer & Digit   \\ \hline
        t-SNE                          & 0.5586 & 0.6530 & 1.8637 & 8.1966  \\ \hline
        PCA                            & 0.0057 & 0.0005 & 0.0062 & 0.0097  \\ \hline
        vanilla FS RANDOM              & 3.4046 & 0.7954 & 1.1158 & 2.6178  \\ \hline
        vanilla FS PCA                 & 3.1948 & 2.4821 & 1.1870 & 1.8550  \\ \hline
        vanilla FS TSNE                & 1.5389 & 1.1543 & 1.9115 & 6.8993  \\ \hline \hline
        DimenFix RANDOM                   & 3.7058 & 0.8299 & 1.0899 & 1.3676  \\ \hline
        DimenFix PCA                      & 2.9507 & 0.9476 & 0.4534 & 2.7260  \\ \hline
        DimenFix TSNE                     & 3.9838 & 1.7048 & 2.8627 & 6.7649  \\ \hline
        DimenFix norm\_mov\_range RANDOM  & 3.6864 & 0.8045 & 1.2872 & 4.6592  \\ \hline
        DimenFix norm\_mov\_range PCA     & 2.9232 & 1.0935 & 2.1563 & 4.8718  \\ \hline
        DimenFix norm\_mov\_range TSNE    & 5.5858 & 2.7024 & 2.3901 & 6.1028  \\ \hline
        DimenFix gauss\_mov\_range RANDOM & 2.4573 & 1.4672 & 0.5157 & 1.4099  \\ \hline
        DimenFix gauss\_mov\_range PCA    & 3.2706 & 0.5242 & 0.7446 & 2.2404  \\ \hline
        DimenFix gauss\_mov\_range TSNE   & 4.1499 & 1.4890 & 1.9189 & 8.8643  \\ \hline
    \end{tabular}
    \caption{Running time (second): Both t-SNE and Force Scheme based methods require a process of creating embedding space. As a result, PCA gives the shortest running time for all four experimental datasets. However, when comparing DimenFix and vanilla Force Scheme, we can see no significant extra running time added as DimenFix is applied. }
    \label{tab:running_time}
\end{table*}

\subsection{Kruskal's Stress} \label{sec:kruskal_score}

Unlike classification tasks, we do not have wildly accepted mathematical metrics for dimensionality reduction results. Neither PCA nor t-SNE implemented in the scikit-learn library has such a built-in metric~\cite{pedregosa2011scikit}. Here we propose to use a modified version of Kruskal's stress~\cite{kruskal1964nonmetric}, which was originally used for measuring Multidimensional scaling(MDS), to measure the quality of projections created by DimenFix and other baselines.

The formula of Kruskal's stress is defined as 

\begin{equation} \label{sec:kruskal_stress}
S = \sqrt{\frac{\sum (d_{ij} - \delta_{ij})^2}{\sum d_{ij}^2}}
\end{equation} 
\noindent 
where $d$ is the distances between each pair of data points in the original dataset, and $\delta$ is the distances (again, in this paper, we are using euclidean distance) between each pair of data points in the dimensionality reduced dataset or the projection. The range of Kruskal's stress is $[0, 1]$, where closer to $0$ means better the projection~\cite{kruskal1964nonmetric}. 

Here we determine $d$ by calculating the triangular distance matrix using the original dataset after a pre-process of scaling. More specifically, we scale the original datasets to a certain range $(n, m)$ where $n$ and $m$ can be defined by the user. This pre-processing is necessary because we are fixing an arbitrary feature on one axis. If the dataset is not scaled and the original scale of the fixed feature has a highly different range than the majority of the features, the resulting dataset might be weird. We then use the pre-processed dataset to create a triangular distance matrix, which will be used later to get $d$. 

After getting the projection, before calculating $\delta$, we again scale the projection with the same range $(n, m)$. This is to eliminate any unnecessary influence brought by scaling differences before and after the projection and only to compare the distances of each pair of data points with the same scale. 

Here the projections we created are again three-dimensional for both baselines and DimenFix. Force Scheme (both vanilla Force Scheme and DimenFix) performs better than PCA and t-SNE, as shown in Table~\ref{tab:kruskal_stress}. This is because of the nature of the Force Scheme, which uses the distances among data points to create projections. It is the distance that Force Scheme is trying to reduce. Hence, Kruskal's stress is calculated based on distances. 

Similar to running time, DimenFix does not increase Kruskal's stress significantly compared with the vanilla Force Scheme (see Section~\ref{sec:running_time}).

\begin{table*}[!ht]
    \centering
    \begin{tabular}{|l|l|l|l|l|}
    \hline
        ~                              & Iris            & Wine            & Cancer          & Digit            \\ \hline
        t-SNE                          & 0.4691          & 0.5612          & 0.6690          & 0.8306           \\ \hline
        PCA                            & 0.3153          & 0.6204          & 0.8228          & 0.8410           \\ \hline
        vanilla FS RANDOM              & \textbf{0.1597} & \textbf{0.4863} & 0.6793          & 0.8299           \\ \hline
        vanilla FS PCA                 & 0.2329          & 0.4975          & 0.6713          & 0.8249           \\ \hline
        vanilla FS TSNE                & 0.1866          & 0.4977          & 0.6859          & 0.8283           \\ \hline \hline
        DimenFix RANDOM                   & \textbf{0.1640} & 0.5047          & 0.6325          & 0.8001           \\ \hline
        DimenFix PCA                      & 0.1995          & \textbf{0.4871} & \textbf{0.6141} & \textbf{0.7982}  \\ \hline
        DimenFix TSNE                     & 0.2017          & 0.4974          & \textbf{0.6218} & \textbf{0.7935}  \\ \hline
        DimenFix norm\_mov\_range RANDOM  & 0.2455          & 0.5031          & 0.6433          & 0.8201           \\ \hline
        DimenFix norm\_mov\_range PCA     & 0.2001          & \textbf{0.4867} & 0.6450          & 0.8214           \\ \hline
        DimenFix  norm\_mov\_range TSNE   & 0.2582          & \textbf{0.4880} & \textbf{0.6169} & 0.8130           \\ \hline
        DimenFix gauss\_mov\_range RANDOM & 0.2816          & 0.5073          & 0.6479          & 0.8091           \\ \hline
        DimenFix gauss\_mov\_range PCA    & 0.2113          & 0.4948          & 0.6471          & 0.8150           \\ \hline
        DimenFix gauss\_mov\_range TSNE   & 0.2704          & 0.4932          & 0.6359          & 0.8088           \\ \hline
    \end{tabular}
    \caption{Kruskal's Stress: This score is designed to measure the similarity and dissimilarity between the raw dataset and a projection using distance. Since t-SNE and PCA do not use distance as a reference for creating the projection, it is not surprising that Force Scheme based methods give a better score. Moreover, DimenFix does not significantly increase the stress score compared with vanilla Force Scheme}
    \label{tab:kruskal_stress}
\end{table*}

\subsection{Classification Performance} \label{sec:classification_performance}

In this subsection, we measure the performance of DimenFix by comparing the classification accuracy of the result projection to baselines. For this experiment, in addition to the t-SNE, PCA, and vanilla Force Scheme, we record the classification accuracy of the whole dataset. The projections created by baselines and DimenFix are still three-dimensional. 

To experiment, we again created three-dimensional projections. We randomly sampled $70\%$ of each experimental dataset as the training set. The remaining $30\%$ are used as a test set. The indexes used to sample training and test set are identical to minimize any influence brought by the difference of data points used within the same dataset (the column-wise results of Table~\ref{tab:classifications}). The classifiers we use are built-in methods in the scikit-learn library~\cite{pedregosa2011scikit}. We choose Random Forest Classifier, Decision Tree Classifier, and Logistic Regression. The starting projection mode we are using for DimenFix here is t-SNE. In the classification results, while the bolded values are the best performances, we are not bolding the results of raw datasets which passed all dimensions to the classifiers (see Table~\ref{tab:classifications}). 

There are several conclusions we can get from the classification performance.

Firstly, we see a difference in the accuracy of the raw dataset, t-SNE projection, PCA projection, and vanilla Force Scheme. These differences come from the fact that the different algorithms (loss functions defined by t-SNE and vanilla Force Scheme or the principle component calculated by PCA) used by t-SNE, PCA, and vanilla Force Scheme change the relationship among points in different ways. In most cases, the raw high-dimensional dataset, which is more informative than the projections, usually has a better classification performance. 

Occasionally, suppose the features regarding which points had changed in their relationship with each other are not decisive for a specific classifier to make the decision. In that case, the projection may give a slightly better classification accuracy than the raw dataset for that particular classifier (e.g., classification by decision tree classifier on the projection of Digit dataset created by t-SNE). This result entirely depends on the dataset and the classifier and is out of the user's control.

If we compare DimenFix with vanilla Force Scheme, on the other hand, the classification accuracy depends on how important the feature fixed is. For example, in the Wine dataset, all three classifiers perform better on the projections created by DimenFix than by a vanilla Force Scheme. Moreover, the moving-in-range modes of DimenFix give a slightly worse performance than the normal mode (strictly fixed mode) in all three classifiers. That is normal DimenFix $>$ moving-in-range DimenFix $>$ vanilla Force Scheme. This means the feature selected to fix ("alkalinity of ash") is very important. We can also see a similar phenomenon from the results of the Breast Cancer dataset. However, the difference in Breast Cancer results between DimenFix and vanilla Force Scheme is less significant. This means the chosen feature ("worst concave points") has a significant influence on the class labels but is not as important as "alkalinity of ash" for the Wine dataset. On the contrary, the features selected in the Iris and Digit datasets are less important than those in the other two.

To conclude, for a particular dataset, if a critical feature exists, applying DimenFix on that feature should give a better projection. On the other hand, by analyzing the classification result of the DimenFix protection, we should also get a sense of how important the fixed feature is to the corresponding dataset.

\begin{table*}[!ht]
    \centering
    \begin{tabular}{|l|l|l|l|l|}
    \hline
        ~                  & Iris (DT)       & Wine (DT)       & Cancer (DT)     & Digit (DT)       \\ \hline
        raw dataset        & 1.0000          & 0.9597          & 0.9824          & 0.9515           \\ \hline \hline \hline
        t-SNE              & 0.9619          & 0.8790          & 0.9774          & \textbf{0.9928}  \\ \hline
        PCA                & \textbf{1.0000} & 0.9274          & 0.9598          & 0.9189           \\ \hline
        vanilla FS         & \textbf{1.0000} & 0.8952          & 0.9774          & 0.9101           \\ \hline \hline
        DimenFix              & 0.9905          & \textbf{0.9919} & 0.9799          & 0.9101           \\ \hline
        DimenFix norm\_range  & 0.9905          & 0.9758          & 0.9749          & 0.9109           \\ \hline
        DimenFix gauss\_range & \textbf{1.0000} & \textbf{0.9919} & \textbf{0.9824} & 0.9045           \\ \hline
    \hline \hline \hline
        ~                  & Iris (RF)       & Wine (RF)       & Cancer (RF)     & Digit (RF)       \\ \hline
        raw dataset        & 1.0000          & 1.0000          & 0.9950          & 0.9936           \\ \hline \hline \hline
        t-SNE              & 0.9714          & 0.9032          & 0.9724          & \textbf{0.9976}  \\ \hline
        PCA                & \textbf{1.0000} & 0.9597          & 0.9749          & 0.9204           \\ \hline
        vanilla FS         & \textbf{1.0000} & 0.9113          & 0.9724          & 0.9324           \\ \hline \hline
        DimenFix              & \textbf{1.0000} & \textbf{0.9919} & 0.9849          & 0.9189           \\ \hline
        DimenFix norm\_range  & \textbf{1.0000} & 0.9839          & 0.9849          & 0.9300           \\ \hline
        DimenFix gauss\_range & \textbf{1.0000} & \textbf{0.9919} & \textbf{0.9950} & 0.9212           \\ \hline
    \hline \hline \hline
        ~                  & Iris (LR)       & Wine (LR)       & Cancer (LR)     & Digit  (LR)      \\ \hline 
        raw dataset        & 0.9810          & 0.9597          & 0.9447          & 0.9889           \\ \hline \hline \hline
        t-SNE              & 0.8667          & 0.6613          & 0.9095          & \textbf{0.9642}  \\ \hline
        PCA                & \textbf{0.9810} & 0.8306          & 0.9347          & 0.7049           \\ \hline
        vanilla FS         & \textbf{0.9810} & 0.7097          & 0.9146          & 0.6388           \\ \hline \hline
        DimenFix              & 0.9429          & 0.9597          & 0.9347          & 0.6874           \\ \hline 
        DimenFix norm\_range  & 0.8952          & 0.9194          & \textbf{0.9372} & 0.6794           \\ \hline
        DimenFix gauss\_range & 0.9048          & \textbf{0.9758} & 0.9347          & 0.6698           \\ \hline
    \end{tabular}
    \caption{Classification Accuracy: Decision Tree (DT) vs Random Forest (RF) vs Logistic Regression (LR). As agreed by all classifiers, the Wine and Cancer datasets have a very important feature, and as the relationship defined by this feature is preserved, the classification accuracy significantly improved}
    \label{tab:classifications}
\end{table*}

\subsection{Visualization} \label{sec:visualization}

\begin{figure*}
    \centering
    \includegraphics[width=\textwidth]{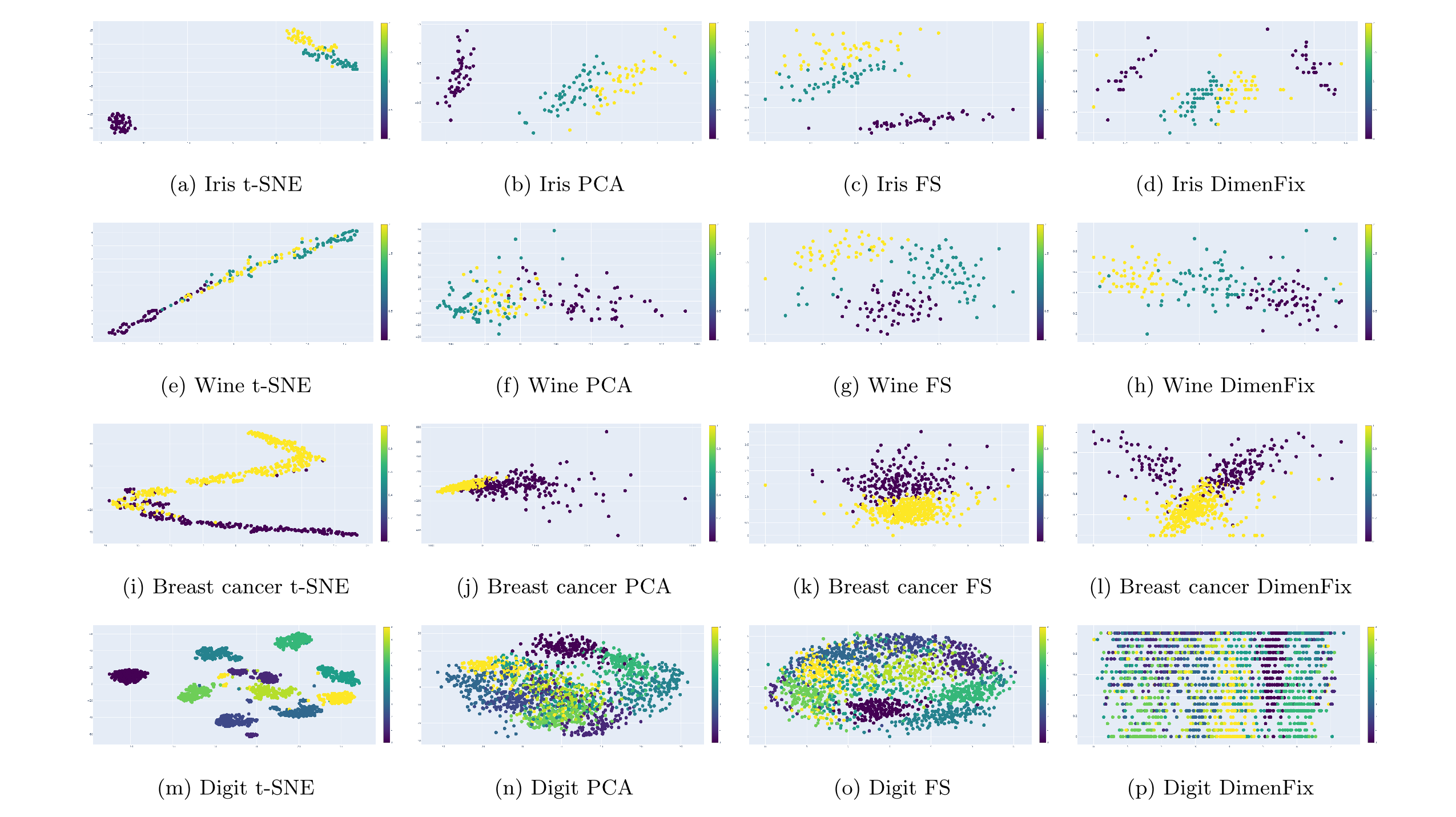}
    \caption{2D visualizations: The projections created by baselines (left-most three columns) only show the relationship of data points regarding the whole dataset. However, in the projections created by DimenFix (rightmost column), the $y$ axis shows the raw value of a particular feature, while the $x$ axis shows the relationship of data points regarding the whole dataset and the particular feature. If the given feature is not significantly more important than other features, a nicely separated visual cluster might not be formed (e.g., Digit dataset).}
    \label{fig:2d_visualization}
\end{figure*}

For visualization, we used Plotly~\cite{plotly} library to visualize the projections. Since a three-dimensional figure cannot be fully printed out in a research paper, to demonstrate the visualization ability better, we this time created two-dimensional projections for the baselines (t-SNE, PCA, vanilla Force Scheme) and DimenFix. The $y$ axis represents the feature to be fixed, while the colors represent the classes (in Fig.~\ref{fig:2d_visualization}).

We have several observations from the projections.

Firstly, as the $y$ axis represents the fixed feature, we can see the class distribution on the fixed feature. For example, in the Breast Cancer dataset, we can see that the two classes have a different distribution on the fixed axis "worst concave points" (Fig.~\ref{fig:2d_visualization} (l)) One class is concentrated on the top part, while another class is concentrated on the bottom of the figure.

Secondly, the $x$ axis is also very informative. The algorithm of Force Scheme is trying to push different points farther away from each other~\cite{tejada2003improved}, but here $y$ axis is fixed. Different points will then move in opposite directions on the $x$ axis. As a result, we can observe that the data point distribution, calculated using all features in the raw dataset, is visualized regarding the fixed feature. This is especially obvious for the Iris and Breast Cancer datasets. We can see that for both datasets, one of the classes formed two obvious subsets within the fixed feature (Fig.~\ref{fig:2d_visualization} (d) and Fig.~\ref{fig:2d_visualization} (l)). 

In short, using DimenFix, we are not just trying to create visual clusters, if there are any of them in the raw dataset, but are trying to reproduce some more detailed and customized relationships among points from a given dataset. The task of visual clustering can be well done by different dimensionality reduction methods, but the results we described in this section can only be observed by DimenFix.

\section{Conclusion} \label{sec:conclusion}

In this paper, we presented a novel meta-method, DimenFix, which is built upon any gradient-descent-based base dimensionality reduction method and allows its user to influence the change in relationships, where the relationships are defined by the loss function of the base dimensionality method between each pair of points in a dataset. 

Unlike normal dimensionality reduction methods, the aim of DimenFix is not only to form visual clusters for datasets containing multiple classes while treating the dataset features uniformly. With DimenFix, different goals can be achieved, like (a) better preserving a particular feature of the dataset while simultaneously reducing the dimensionality and (b) understanding the data distribution with respect to a specific feature specified by the user.

DimenFix maintains an equally good running time and Kruskal's Stress score compared with vanilla Force Scheme and other baselines while enabling a novel feature preservation task, where the feature is selected by the user. For those datasets containing a critical feature, DimenFix gives a better projection than the baselines, as demonstrated by the higher classification accuracies. When the projection is visualized, DimenFix also provides the users with more information related to the particular feature, which is not revealed by any baselines.

\section{Future Work} \label{sec:future_work}

Several future research directions can be pursued.

Firstly, we have tested DimenFix using Force Scheme as the base dimensionality reduction method~(Section~\ref{sec:results}). The methodology for applying DimenFix to other base methods is identical to what we did for Force Scheme (the URL of the source code can be found in Section~\ref{sec:introduction}), but the result might vary significantly. DimenFix is applied to the gradient-descent-like process. In turn, the base method's performance is affected by the loss function, and therefore it may vary from method to method. For example, in Force Scheme, this loss function is defined by the Euclidean distance between each pair of points, while for t-SNE, this loss function is defined by the probability distribution among points. Hence, even though the inference made by DimenFix on any base method is identical, we might see different results. In the future, we also plan to test the performance of DimenFix based on other gradient-descent-based dimensionality reduction methods, like t-SNE or UMAP.

Secondly, DimenFix, so far, is focused on 2- and 3-dimensional projection. If visualization is not necessary, and if more than one feature is more important than the others in a dataset, a higher-dimensional projection will become helpful. 

Thirdly, DimenFix can only be applied to an entire dataset. Sometimes, the importance of some data points within a dataset might be higher than the rest. In this case, it would be helpful for a user to define a part of a dataset to be fixed or define the moving range point by point. A user interface and an interactive process would help achieve this goal. Ideally, in such a user interface, a user should be able to interact with the projection-creating process by understanding and changing the relationships among points shown by the current projection.

\bibliographystyle{eg-alpha-doi}
\bibliography{references.bib}

\end{document}